\documentclass{article} 
\usepackage{iclr2024_conference,times}

\usepackage{amsmath,amsfonts,bm}

\def\eqref#1{equation~\ref{#1}}

\def\1{\bm{1}}

\DeclareMathAlphabet{\mathsfit}{\encodingdefault}{\sfdefault}{m}{sl}
\SetMathAlphabet{\mathsfit}{bold}{\encodingdefault}{\sfdefault}{bx}{n}

\usepackage{hyperref}
\usepackage{url}
\usepackage{microtype}
\usepackage[final]{graphicx}
\usepackage{subcaption}
\usepackage{float}
\usepackage{booktabs}
\usepackage{dblfloatfix}
\usepackage{chngcntr}
\usepackage{amsmath,amssymb}

\title{Sharpness-Aware Minimization (SAM) Improves \\ Classification Accuracy of Bacterial Raman \\ Spectral Data Enabling Portable Diagnostics}

\author{Kaitlin Zareno$^*$, Jarett Dewbury$^*$, Loza F. Tadesse\\
Department of Mechanical Engineering\\
Massachusetts Institute of Technology\\
Cambridge, MA 02139, USA \\
\texttt{\{kzareno,jdewbury,lozat\}@mit.edu} \\
\And
Siamak K. Sorooshyari\\
Department of Statistics\\
Stanford University\\
Palo Alto, CA 94305, USA\\
\texttt{siamak@stanford.edu}
\AND
Hossein Mobahi \\
Google Research\\
Mountain View, CA 94043, USA\\
\texttt{hmobahi@google.com}
}

\iclrfinalcopy

\begin{document}

\maketitle

\begin{abstract}
Antimicrobial resistance is expected to claim 10 million lives per year by 2050, and resource-limited regions are most affected. Raman spectroscopy is a novel pathogen diagnostic approach promising rapid and portable antibiotic resistance testing within a few hours, compared to days when using gold standard methods. However, current algorithms for Raman spectra analysis 1) are unable to generalize well on limited datasets across diverse patient populations and 2) require increased complexity due to the necessity of non-trivial pre-processing steps, such as feature extraction, which are essential to mitigate the low-quality nature of Raman spectral data. In this work, we address these limitations using Sharpness-Aware Minimization (SAM) to enhance model generalization across a diverse array of hyperparameters in clinical bacterial isolate classification tasks. We demonstrate that SAM achieves accuracy improvements of up to $10.5\%$ on a single split, and an increase in average accuracy of $2.7\%$ across all splits in spectral classification tasks over the traditional optimizer, Adam. These results display the capability of SAM to advance the clinical application of AI-powered Raman spectroscopy tools. 
Code is available at: \url{https://github.com/Tadesse-Lab/SAM-Raman-Diagnostics}

\end{abstract}

\section{Introduction}
Antimicrobial resistance is the second biggest global health threat and is expected to surpass cancer by 2050 \citep{deKraker2016} as the second leading cause of death in the United States. According to the Centers for Disease Control and Prevention (CDC), nearly $50\%$ of all outpatient antibiotic usage is improper, including unnecessary use and inappropriate selection, dosing, and duration of antibiotic treatments \citep{Centers_for_Disease_Control_and_Prevention_CDC2011-fn,Pichichero2002-ts,Shapiro2014-yn}. Altogether, such usage compromises effectiveness and contributes significantly to antimicrobial resistance. In the context of this escalating challenge, it is crucial to recognize that certain regions, such as developing nations or conflict zones, where access to quality hospital-level clinical care is limited, are most vulnerable. Thus, a rapid, reliable, and compact solution for bacterial infection diagnosis is needed to provide timely and accurate treatment. A promising tool is Raman spectroscopy, an optical technique that captures inelastically scattered light acting as a fingerprint for precise mapping to specific pathogen types. Within seconds, this method can identify not only the primary pathogen but also its mutated antibiotic resistant variants \citep{das2011vibspec,Garcia-Rico2018-nm,Pieczonka2008-xg}.

In this work, we address the intrinsic data quality challenges of working with Raman spectral data by utilizing Sharpness-Aware Minimization (SAM) as an optimizer for Raman spectral data analysis. Our main contribution is as follows: we demonstrate that SAM is an effective optimizer for clinical Raman spectroscopy-based pathogen classification tasks, achieving an increase in accuracy of up to $10.5\%$ on a single split, and an increase in average accuracy of $2.7\%$ across all splits when compared to a standard ResNet architecture with Adam. The results demonstrate accuracy and generalizability in translating rapid on-device pathogen diagnostics for patients in need.

\section{Background and Related Work}
Although Raman spectroscopy is a promising tool for portable and rapid infectious disease diagnostics, Raman spectral data tends to be complex, high-dimensional, and noisy. Due to intrinsic limitations in the quality of Raman spectral data, current ML-based (machine learning) approaches in Raman spectroscopy often encounter 1) increased complexity due to the necessity of extra enhancing steps, such as feature extraction \citep{Gautam2015-lj,Pelletier2003-jt}, and 2) poor generalization due to limited datasets and small sample sizes, which have high variability both within individual patient samples and across patient populations \citep{Luo2022-dl}. These data challenges limit the clinical translation of portable Raman spectroscopy-based diagnostics.

Deep learning offers the advantage of extracting non-linearities in the data without the need for non-trivial pre-processing \citep{Liu2017-in},  allowing for improved classification and more rapid processing in clinical translation. 1D convolutional neural networks (CNNs) and ResNets are the most commonly applied models in Raman classification tasks, and certain architectures (Figure \ref{fig:figure_1}) have achieved accuracies of up to $99\%$ in classifying pathogen types and antibiotic resistance using Raman spectra \citep{Ho2019-et,Ogunlade2023-ro}. However, prior ML classification tasks have required multiple samples to make a majority class prediction, and since sampling is a time-consuming process, it is less desirable for clinical translation where rapid assessment is needed. 

SAM offers a potential solution to the generalization challenges that come with the application of traditional ML approaches to small datasets. Unlike previous optimization methods that may develop sharp minima and have poor generalization, SAM is an optimization technique that improves generalization by minimizing both the loss value and loss sharpness simultaneously \citep{Foret2020-gd}. The ability to effectively generalize to unseen data makes SAM a useful tool for addressing the limited data issue commonly encountered in deep learning inference tasks with Raman spectral data. Despite its success in healthcare-related tasks \citep{Song2022-ok,Anand2022} and its promise towards advancing generalization in tasks with data quality concerns, SAM has not yet been applied to Raman spectral data.  In this work, we address this gap by using SAM to enhance model generalization across a diverse array of hyperparameters in patient-derived bacterial classification tasks.

\section{Methods}
\subsection{Sampling Method:}
Clinical significance was preserved through subject-level sampling by excluding complete patient samples from either the train or test set. This approach ensured that our model had not seen any spectra from the clinical isolates in the test set during training, maintaining the integrity of the simulation on a subject-level. We utilized this method in our clinical evaluation pipeline as described below.

To demonstrate our algorithm’s performance, we used publicly available spectral data from a leading Raman spectroscopy and CNN-based work by \cite{Ho2019-et}. We obtained clinical bacterial isolate spectral data spanning the five most common first-line antibiotic treatment groups. This data was collected in groups of two distinct biological replicates. The dataset comprises 50 distinct clinical isolates collected from individual patients, categorized into five bacterial pathogen classes. It includes an equal distribution of five patient sources per infection type. To assess the efficacy of SAM with a ResNet architecture for Raman we performed random stratified splitting on this clinical spectral dataset.

To demonstrate clinically relevant evaluation, we began by randomly assigning one patient per infection type from each of the two biological replicates to the test set (Figure \ref{fig:appendix_1}). The remaining four patients in each type were utilized for training where they were split 90/10 into train and validation sets. We then evaluated each model on the test dataset which contains independent clinical data that has not been seen in the training pipeline. We repeated this process 10 times, employing randomized splits for each iteration (Figure \ref{fig:figure_a1b}). This approach was utilized to enhance variation within test sets and introduce a diverse set of data splits. We report classification results on the test dataset across five trials per split. 

\subsection{Architecture Configuration and Select Hyperparameters}
\begin{figure*}[!t]
\captionsetup{skip=0.2\baselineskip, labelfont={normalfont}}
\captionsetup[subfigure]{skip=1pt, singlelinecheck=false}
    \centering
      \centering
      \includegraphics[width=.85\textwidth]{"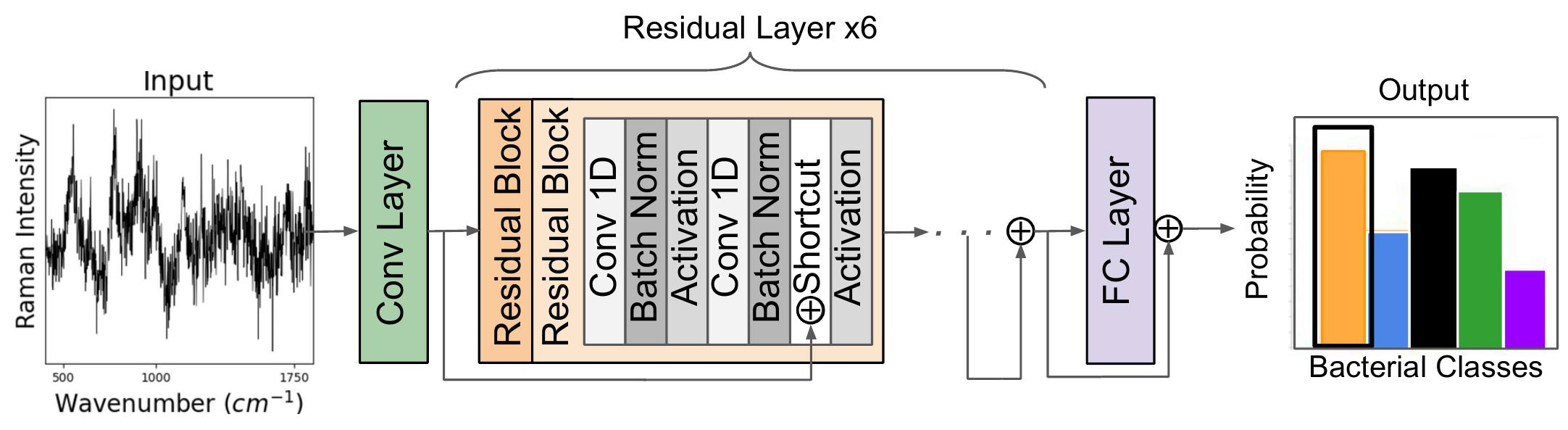"}
      \label{fig:figure_1a}
  \vspace{5px}
\caption{A CNN trained on clinical Raman spectral data is used for a pathogen classification task. Publicly available spectral data and CNN architecture are adapted from \cite{Ho2019-et}. Using a one-dimensional residual network with 6 residual blocks and 25 total convolutional layers, clinical Raman spectra are classified as one of 5 isolates.}
\label{fig:figure_1}
\end{figure*}

For our baseline architecture, we utilized the ResNet described in \cite{Ho2019-et} (Figure \ref{fig:figure_1}) due to the architecture's proven efficacy in bacterial classification tasks. Throughout the course of experimentation, this architecture and the associated hyperparameters were modified in order to optimize model performance. During experimentation, we examined the use of Gaussian Error Linear Unit (GELU) activation \citep{Hendrycks2016-yr}, Rectified Linear Unit (ReLU) activation \citep{Nair2010}, and Scaled Exponential Linear Unit (SELU) activation \citep{Klambauer2017-dd}. We trained all models on a single host having 1 NVIDIA T4 GPU. 

\subsection{SAM Improves Generalization by Finding Flat Minima}
For our baseline optimizer, we used vanilla Adam \citep{Kingma2014-ee}, a common choice in deep learning. While Adam and similar optimizers focus on finding parameters with low loss values, the loss surface geometry (particularly its sharpness around optimal solutions) has been shown to significantly affect model generalization \citep{Keskar2016-zn}. This observation led to the development of Sharpness-Aware Minimization (SAM), which seeks parameters in regions of uniformly low loss \cite{Foret2020-gd}. We utilized SAM as described in \cite{Foret2020-gd} to find parameters $\boldsymbol{w}$ within flat loss basins by optimizing the following objective:

\begin{equation}
\label{sam-loss}
\begin{aligned}
\min_{\boldsymbol{w}} \quad & L^{SAM}_\mathcal{S}(\boldsymbol{w}) + \lambda \|\boldsymbol{w}\|_2^2 \\
\text{where} \quad & L^{SAM}_\mathcal{S}(\boldsymbol{w}) = \max_{\|\boldsymbol{\epsilon}\|_p\leq \rho} L_\mathcal{S}(\boldsymbol{w}+\boldsymbol{\epsilon})
\end{aligned}
\end{equation}

To minimize $L^{SAM}_\mathcal{S}(\boldsymbol{w})$, the inner maximization was linearized and solved analytically so that the gradient $\nabla_{\boldsymbol{w}} L^{SAM}_\mathcal{S}(\boldsymbol{w})$ could be effectively approximated as follows: 

\begin{equation}
\label{eq:sam-gradient}
\nabla_{\boldsymbol{w}} L^{SAM}_\mathcal{S}(\boldsymbol{w}) \approx {\nabla_{\boldsymbol{w}} L_\mathcal{S}(w)|_{\boldsymbol{w}+\hat{\boldsymbol{\epsilon}}(\boldsymbol{w})}}
\end{equation}

The model is more likely to generalize effectively and resist perturbations as the resulting parameter values which exist in neighborhoods with uniformly low loss result in minimal variation in loss relative to the training objective function \citep{Keskar2016-zn}. Thus, by finding regions of uniformly low loss, SAM’s optimization algorithm was shown to improve model generalization performance. This makes it ideal for limited, high-variance Raman spectra derived from clinical samples. For our experiments, we utilized a PyTorch implementation of SAM \footnotemark[1]. 

\footnotetext[1]{\url{https://github.com/davda54/sam}}

\begin{figure*}[!t]
\captionsetup{font={},skip=0.1\baselineskip}
\captionsetup[subfigure]{font={}, skip=1pt, singlelinecheck=false}
\centering
\begin{minipage}[t]{0.43\textwidth}
    \centering
    \subcaption{}
    \includegraphics[width=\textwidth]{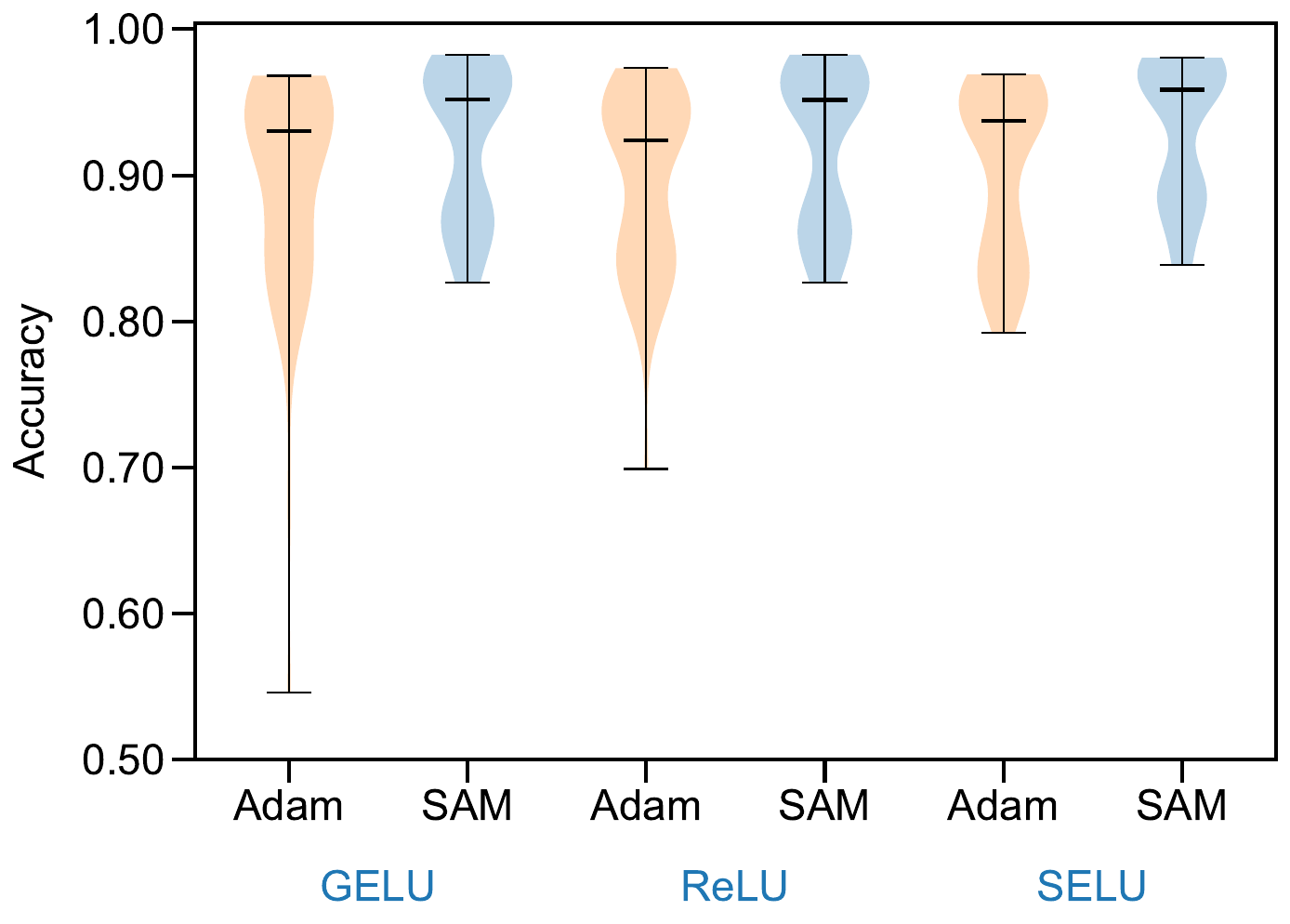}
    \label{fig:2a}
    \subcaption{}
    \vspace{-0.2in}
    \hspace{0.3in}
    \scriptsize
    \begin{tabular}{l|c c c}%
        \toprule
          Optim. & GELU & ReLU & SELU \\
         \midrule
         Adam & $89.3_{8.0}$ & $89.7_{6.1}$ & $90.4_{5.7}$ \\
         SAM & $92.3_{5.2}$ & $92.1_{5.4}$ & $93.1_{4.8}$ \\
        \bottomrule
    \end{tabular}
    \label{fig:2b}
\end{minipage}
\begin{minipage}[t]{0.49\textwidth}
    \centering
    \subcaption{}
    \includegraphics[width=\textwidth]{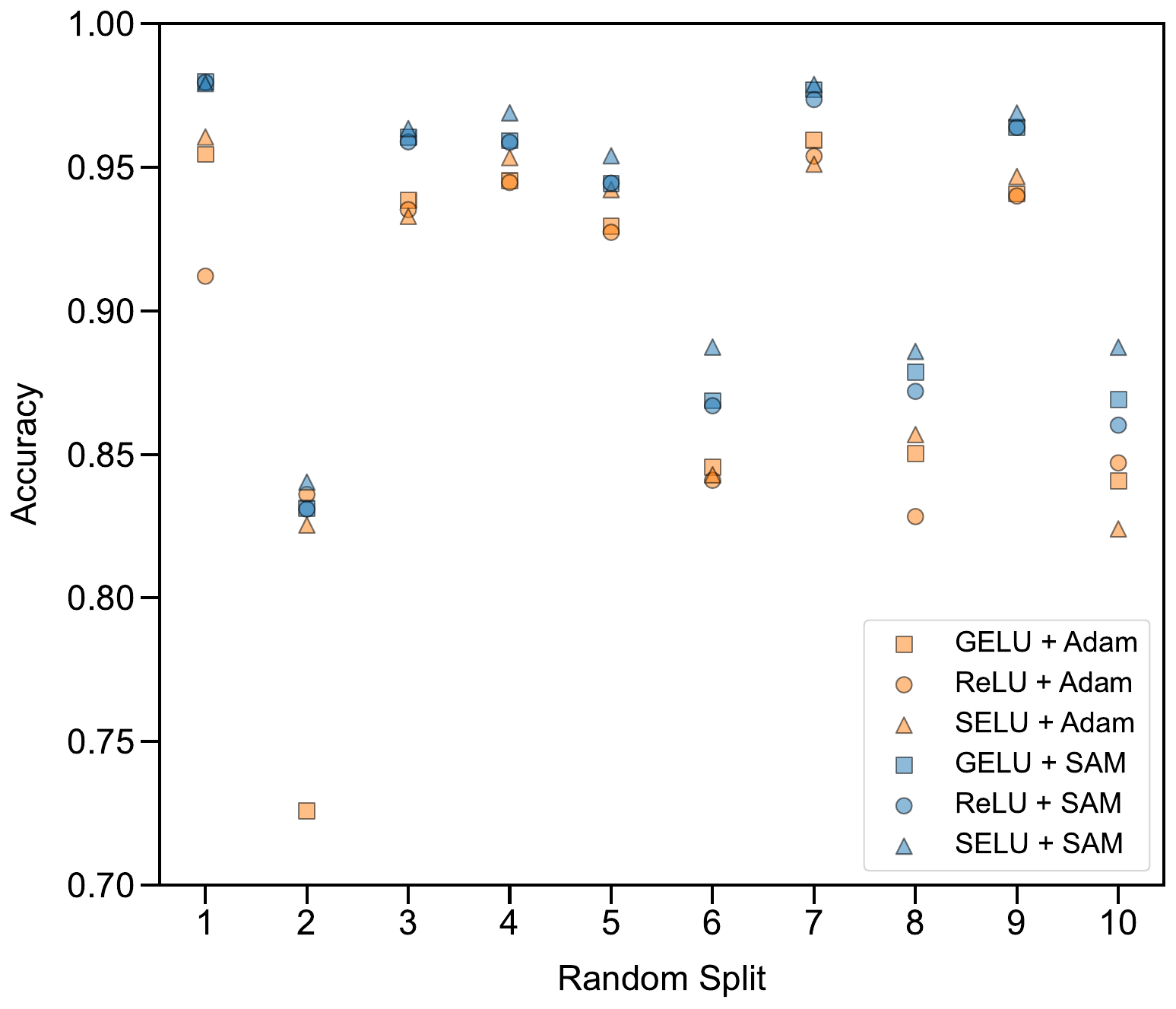}
    \label{fig:2c}
\end{minipage}\vspace{5px}
\caption{Performance of SAM relative to Adam across splits and hyperparameters. (a) A plot visualizing classification accuracies for each model, averaged across all 10 splits. We highlight SAM’s performance, indicating higher accuracy and lower variance across hyperparameters. (b) A table reporting the average classification accuracy of each configuration, showing that SAM increases average accuracy relative to Adam across hyperparameters. SAM with SELU has the best overall performance with an average accuracy of $93.1 \pm 4.8\%$. (c) A summary of results for each individual split demonstrates that ResNet configurations with SAM (blue) tend to outperform those using Adam (orange) across all splits.}
\label{fig:mainfig}
\end{figure*}
\section{Results and Discussion}
\begin{table*}[!b]
\tiny
    \centering
    \caption{Testing set performance ($mean_{std}$ for n=5) across different splits and hyperparameters trained with Adam and SAM optimizers. Training with SAM yields higher classification accuracy across most of the splits and hyperparameters when compared to Adam.} \vspace{5pt}
    
    \resizebox{\linewidth}{!}{
    \begin{tabular}{ll|c c c c c c c c c c}%
        \toprule
        & & \multicolumn{10}{c}{Folds} \\ 
          & Optim. & 1 & 2 & 3 & 4 & 5 & 6 & 7 & 8 & 9 & 10 \\
         \midrule
         GELU & Adam & $95.5_{1.2}$ & $72.6_{10.2}$ & $93.9_{0.7}$ & $94.5_{0.9}$ & $92.9_{0.4}$ & $84.6_{1.2}$ & $96.0_{0.7}$ & $85.0_{2.5}$ & $94.1_{1.1}$ & $84.1_{1.3}$\\
         & SAM & $98.0_{0.2}$ & $83.1_{0.4}$ & $96.0_{0.3}$ & $95.9_{0.1}$ & $94.4_{0.2}$ & $86.9_{0.3}$ & $97.7_{0.1}$ & $87.9_{0.5}$ & $96.4_{0.2}$ & $86.9_{0.3}$\\
         \midrule
        ReLU & Adam & $91.2_{10.7}$ & $83.6_{2.2}$ & $93.5_{0.8}$ & $94.4_{0.6}$ & $92.7_{1.0}$ & $84.1_{1.5}$ & $95.4_{1.6}$ & $82.8_{2.4}$ & $94.0_{2.1}$ & $84.7_{0.8}$\\
         & SAM & $98.0_{0.2}$ & $83.1_{0.4}$ & $95.9_{0.1}$ & $95.9_{0.2}$ & $94.5_{0.4}$ & $86.7_{0.6}$ & $97.4_{0.1}$ & $87.2_{0.2}$ & $96.4_{0.1}$ & $86.0_{0.5}$\\
         \midrule
         SELU & Adam & $96.1_{1.0}$ & $82.5_{1.0}$ & $93.3_{2.2}$ & $95.3_{1.7}$ & $94.2_{0.4}$ & $84.3_{0.8}$ & $95.1_{0.7}$ & $85.7_{2.8}$ & $94.7_{1.4}$ & $82.4_{1.9}$\\
         & SAM & $97.9_{0.1}$ & $84.0_{0.1}$ & $96.4_{0.2}$ & $96.9_{0.2}$ & $95.4_{0.1}$ & $88.7_{0.3}$ & $97.9_{0.2}$ & $88.6_{0.3}$ & $96.9_{0.1}$ & $88.7_{0.2}$\\
    \bottomrule
    \end{tabular}}
    \label{tab:table1}
\end{table*}

Given the implemented ResNet architecture and hyperparameters, we observed that the SAM optimizer increased average accuracy up to $3.0\%$ when compared to Adam, as shown in Figure \ref{fig:2b}. Average performance was determined for each configuration by averaging across all of the 10 splits and 5 trials. Thus, the results reflect the expected model performance and inherent variances within the test dataset. Table \ref{tab:table1} presents a detailed summary of the results across all evaluated splits. As illustrated in Figure \ref{fig:2c}, individual data splits highlight SAM’s performance; when compared to Adam, SAM demonstrated a maximum increase in performance of $10.5\%$ on a single split. SAM consistently outperformed Adam in mean classification accuracy across all splits. For each split, GELU experienced, on average, an increase in accuracy of $3.0\%$ with SAM compared to Adam. Similarly, ReLU and SELU achieved $2.4\%$ and $2.8\%$ increases in average accuracy, respectively. In addition, across all splits, SAM considerably decreased the expected variance in accuracy. When using GELU, ReLU, and SELU, SAM also demonstrated decreases in variance of $79.1\%$, $75.2\%$, and $84.3\%$ on average, respectively, as shown in Table \ref{tab:table1}.

\begin{figure}[!t]
\captionsetup{font={},skip=0.25\baselineskip}
\captionsetup[subfigure]{font={}, skip=1pt, singlelinecheck=false}
\begin{center}
    \begin{subfigure}{0.5\columnwidth}
    \caption{}
    \centerline{\includegraphics[width=\columnwidth]{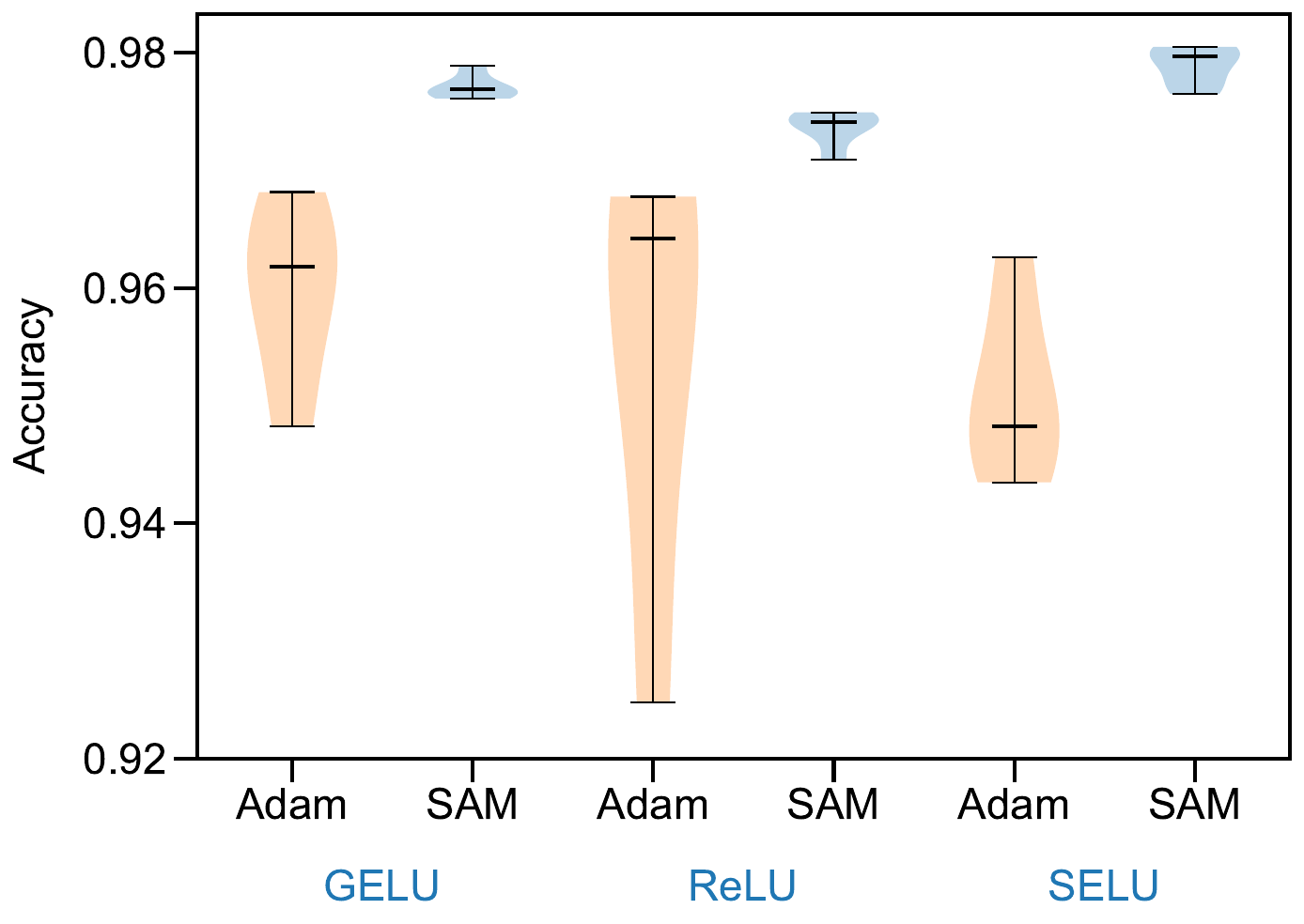}}
    \label{fig:figure_3a}
    \end{subfigure}\hspace{\fill}%
    \begin{subfigure}{0.5\columnwidth}
    \caption{}
    \centerline{\includegraphics[width=\columnwidth]{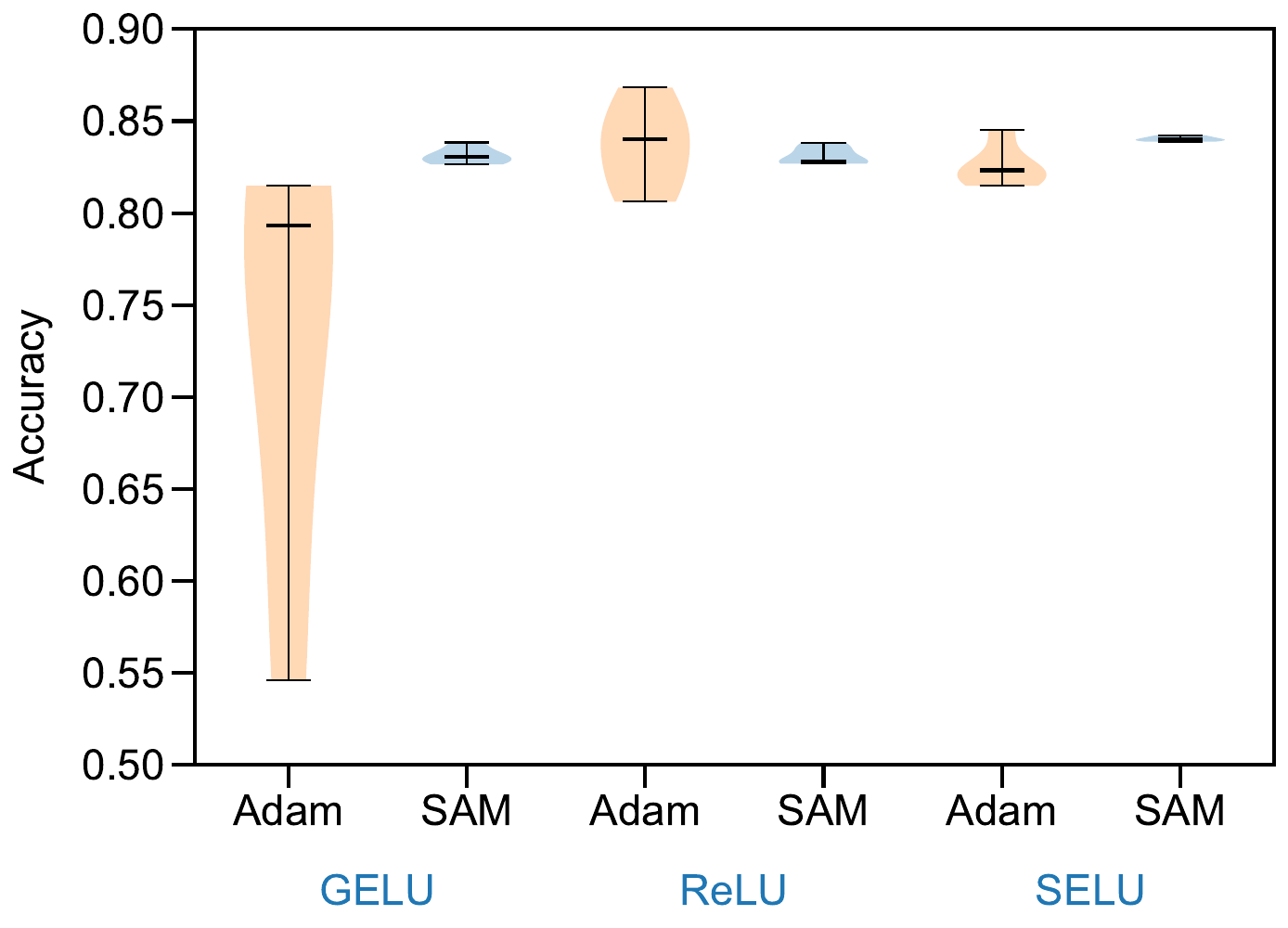}}
    \label{fig:figure_3b}
    \end{subfigure}\hspace{\fill}%
    \vspace{5px}
    \caption{Classification accuracy of each technique on the best and worst performing splits across all hyperparameters. (a) ResNet trained with SAM significantly outperforms Adam across all hyperparameters in the best performing split. (b) Training with SAM has lower variance but no significant performance improvement over Adam in the worst performing split.}
    \label{fig:figure_3}
    \vspace{-11px}
\end{center}
\end{figure}

In addition to looking at the splits as a whole, we observed the best and worst performing splits to better understand how SAM performed at the limits of our test sets (Figure \ref{fig:figure_3}). The best and worst performing splits were determined by observing the mean performance across all splits. In the best performing split (Figure \ref{fig:figure_3a}), we found that SAM significantly outperformed Adam across all hyperparameters, offering both increased accuracy and decreased performance variance. In the worst performing split (Figure \ref{fig:figure_3b}), we observed that SAM provided no significant performance improvement compared to Adam. Figure \ref{fig:appendix_2} further details the performance of SAM and Adam across the remaining splits. Although overlapping regions of standard deviations (Figure \ref{fig:2a}) are observed in average performance across all splits, the lower variance and statistically significant improved accuracy performance in the best performing cases suggest that SAM is positioned to be a more promising approach for clinical Raman spectral classification tasks. The increase in accuracy observed when using SAM as opposed to Adam indicates the promise of SAM for spectral data classification tasks. Furthermore, in clinical classification tasks, consistency in prediction is as crucial as performance itself. SAM exhibited a $2.7\%$ increase in average performance and a $79.5\%$ reduction in average variance, calculated by assessing the mean performance and percent variance per split between each setup using SAM and Adam. This improvement in performance and decreased variance suggest a more stable model to provide greater reliability in prediction for classification tasks where the ability to validate results is limited. 

\section{Conclusion}
In this work, we developed solutions for addressing limitations associated with limited datasets in novel Raman spectroscopy-based rapid infection diagnostics and antibiotic susceptibility testing tools. Compared to the standard utilization of the Adam optimizer for Raman spectral analysis, training with SAM demonstrated an increase in accuracy of up to $10.5\%$ on a single split, an average accuracy improvement of $2.7\%$ across all splits, and considerable reduction in variance across the chosen hyperparameters. Performance with SAM was consistent across 10 distinct splits, showcasing its robustness in processing a variety of patient-derived spectral data. Furthermore, the results indicate a potential to better classify diverse bacterial profiles, suggesting that SAM may provide a more reliable approach for clinical single-spectrum bacterial classification. Our work contributes to the translation of portable Raman spectroscopy-based diagnostics towards the fight against the global health threat of antibiotic resistance.

\textbf{Limitations.} We recognize that further exploration is required to determine the viability of SAM across different datasets and architectures. Furthermore, we acknowledge that the clinical dataset employed in this study may not comprehensively represent patient demographics and may not account for the variance encountered when collecting spectra samples from different Raman instruments. 

\subsection*{Acknowledgments}
We acknowledge support from the MIT-Google Program for Computing Innovation.



\bibliography{iclr2024_conference}
\bibliographystyle{iclr2024_conference}

\newpage

\appendix
\counterwithin{figure}{section}
\section{Appendix}
\subsection{Experimental Methodologies and Results}

\begin{figure*}[h]
\captionsetup{skip=0.25\baselineskip, labelfont={normalfont}}
\captionsetup[subfigure]{skip=1pt, singlelinecheck=false}
    \centering
    \begin{subfigure}{0.64\textwidth}
    \caption{}
      \centering
      \includegraphics[width=\textwidth]{"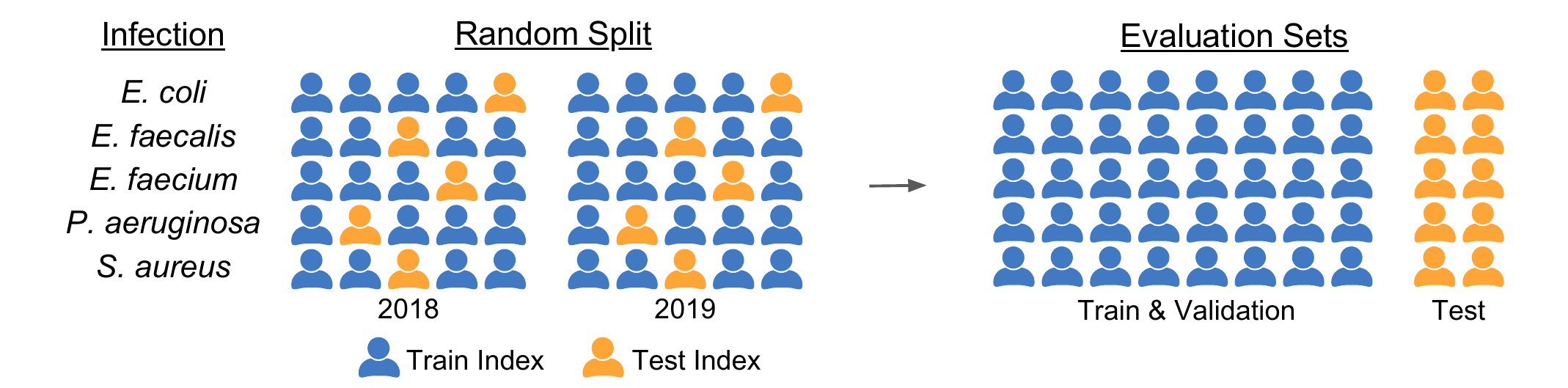"}
      \label{fig:figure_a1a}
    \end{subfigure}\hspace{\fill}%
    \begin{subfigure}{0.64\textwidth}
    \caption{}
      \centering
      \includegraphics[width=\textwidth]{"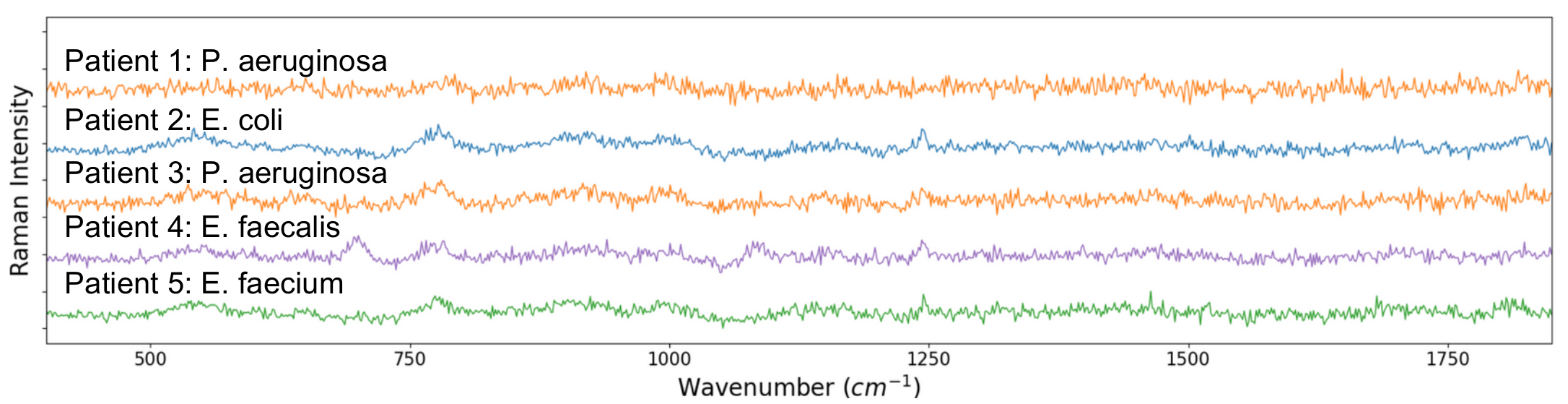"}
      \label{fig:figure_a1b}
    \end{subfigure}\hspace{\fill}%
\caption{A CNN trained on clinical Raman spectral data is utilized for a pathogen classification task. Publicly available spectral data and CNN architecture are adapted from \cite{Ho2019-et}. (a) Two datasets of 5 patients per infection type and 5 species of bacterial infections are merged to create a single dataset of 5 species of bacterial infection and 10 patients per infection type. Each patient is classified into one of the 5 treatment classes where each species corresponds to a different treatment class. Average species identification accuracy using SAM shows an improvement from $89.7 \pm 4.9\%$ to  $92.9 \pm 4.4\%$. (b) Noisy spectra and the intrinsic qualities of Raman spectral data can make it difficult to distinguish between species, highlighting the need for deep learning and SAM to enable rapid analysis of Raman spectral data.}
\label{fig:appendix_1}
\end{figure*}

\begin{figure*}[h]
    \centering
    \begin{subfigure}{.25\textwidth}
      \centering
      \includegraphics[width=\textwidth]{"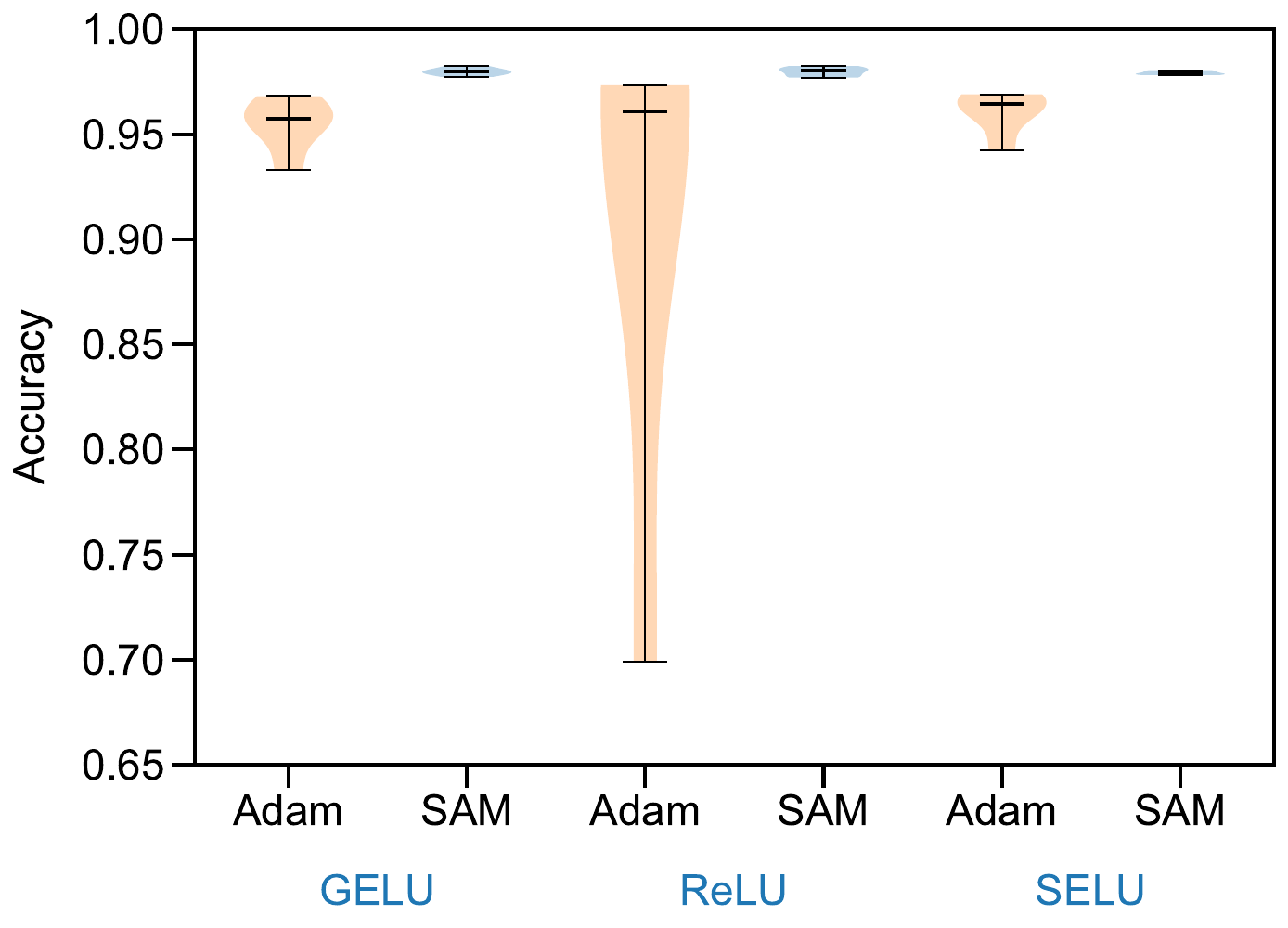"}
      \subcaption{Split 1}
    \end{subfigure}\hspace{\fill}%
    \begin{subfigure}{.25\textwidth}
      \centering
      \includegraphics[width=\textwidth]{"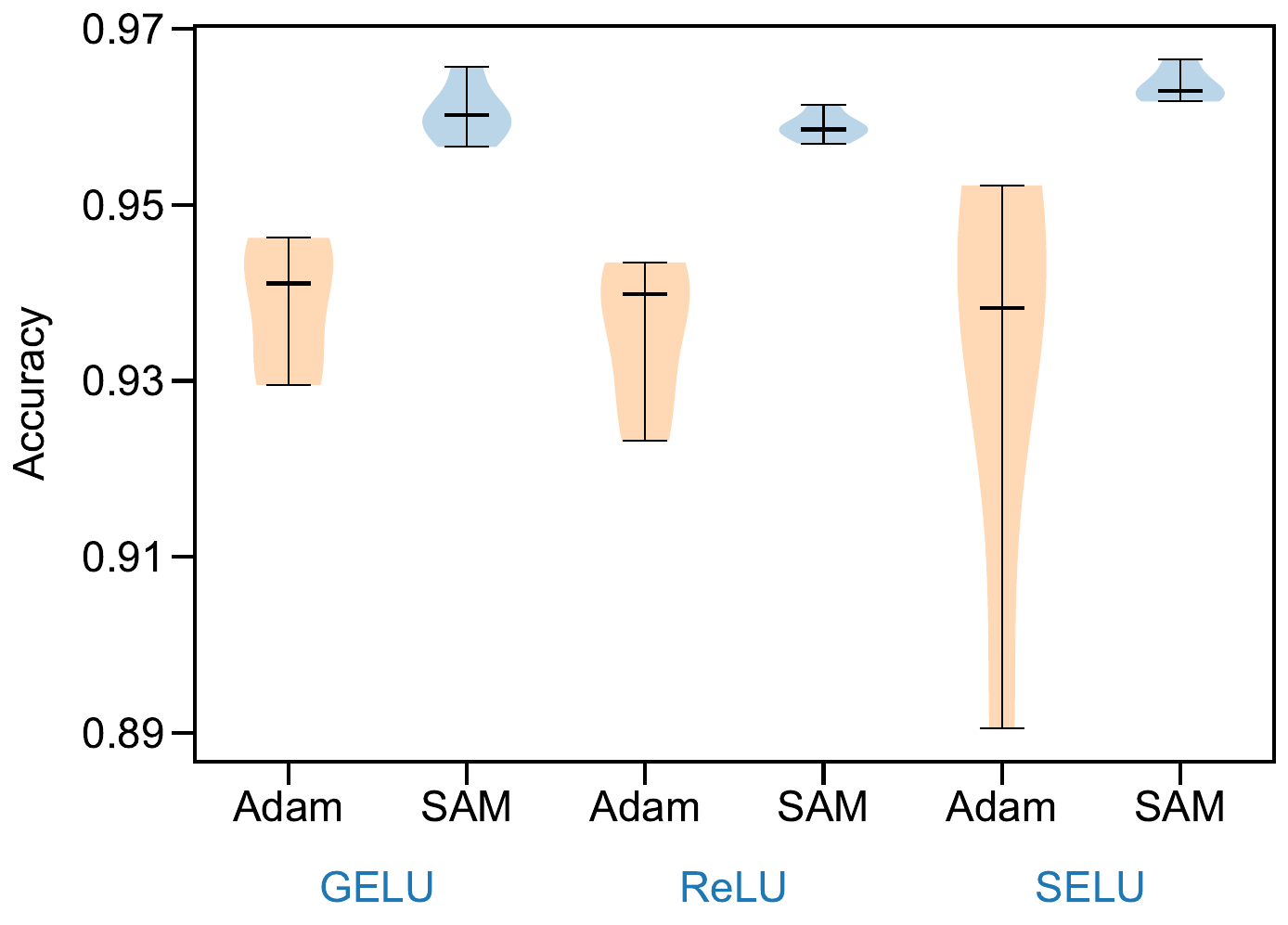"}
      \caption{Split 3}
    \end{subfigure}\hspace{\fill}%
    \begin{subfigure}{.25\textwidth}
      \centering
      \includegraphics[width=\textwidth]{"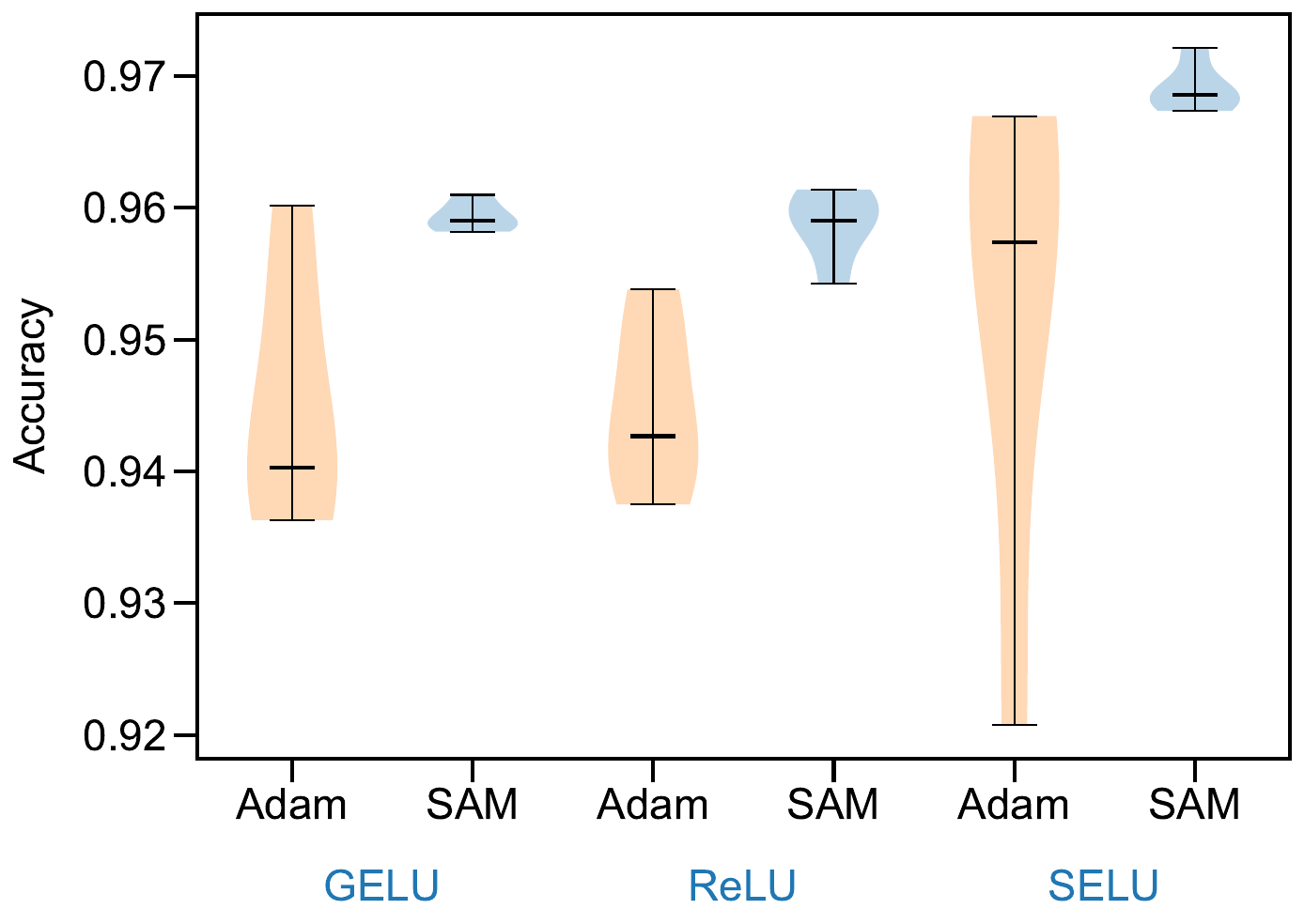"}
      \caption{Split 4}
    \end{subfigure}\hspace{\fill}%
    \bigskip
    \begin{subfigure}{.25\textwidth}
      \centering
      \includegraphics[width=\textwidth]{"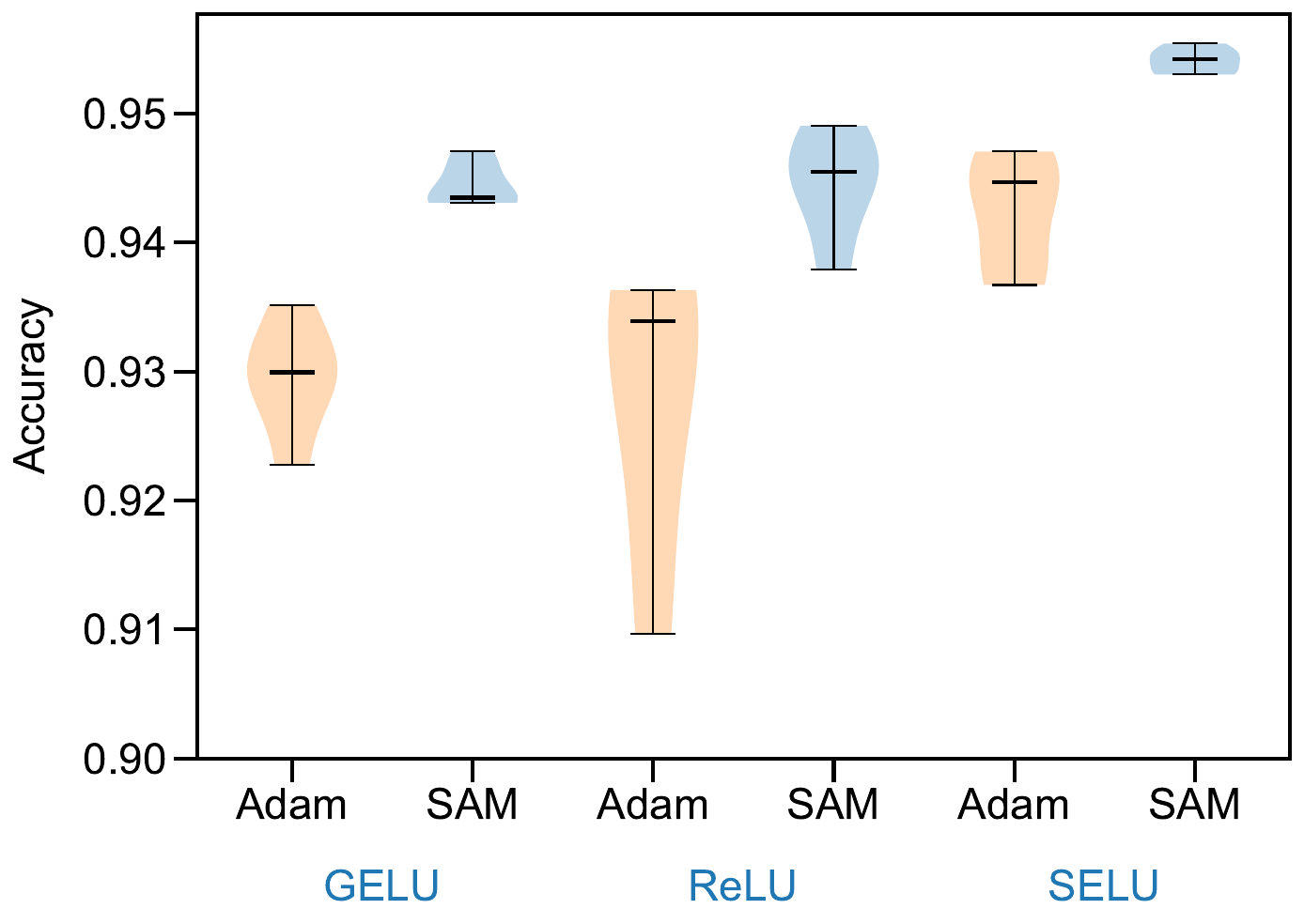"}
      \caption{Split 5}
    \end{subfigure}\hspace{\fill}%
    \begin{subfigure}{.25\textwidth}
      \centering
      \includegraphics[width=\textwidth]{"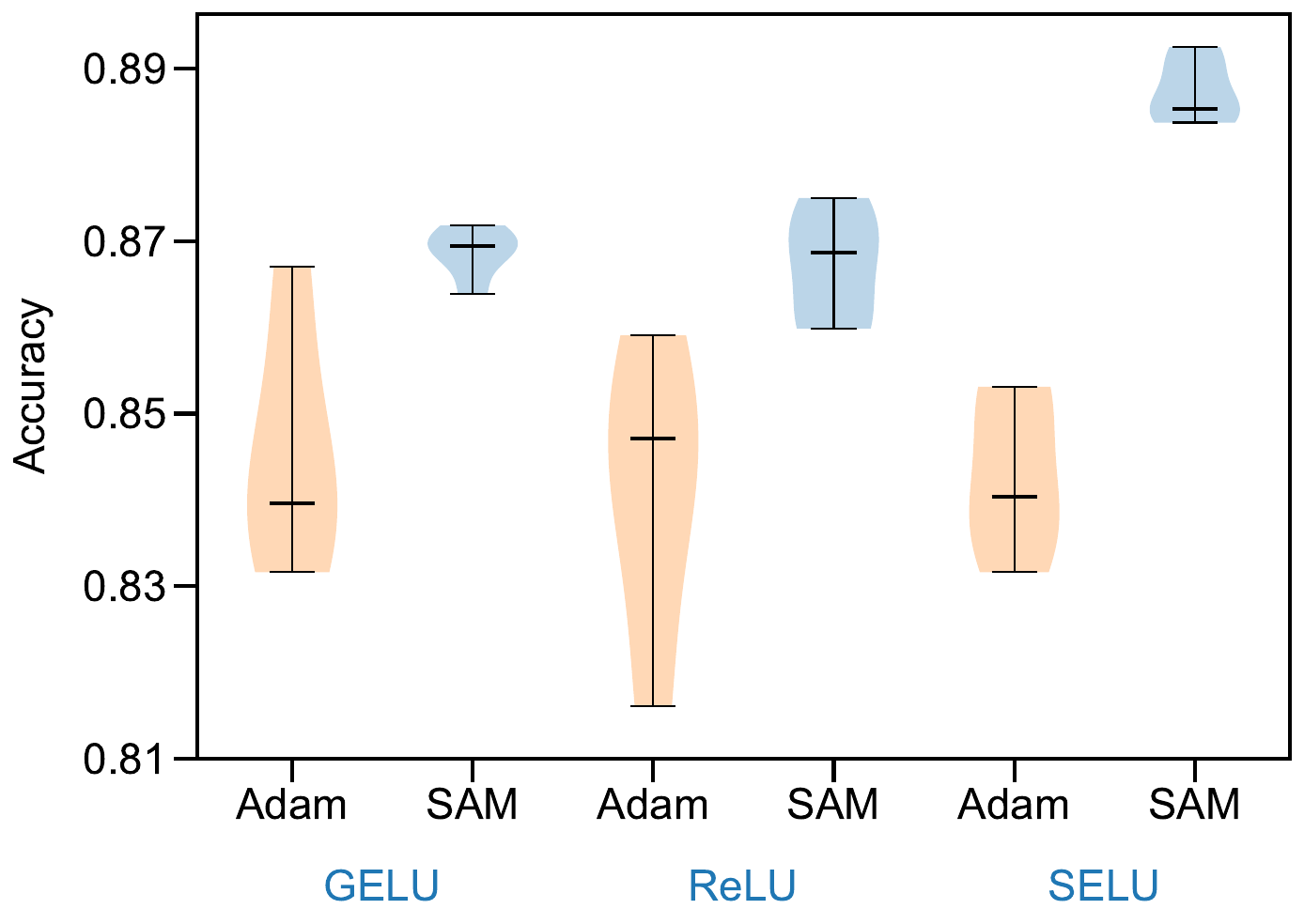"}
      \subcaption{Split 6}
    \end{subfigure}\hspace{\fill}%
    \begin{subfigure}{.25\textwidth}
      \centering
      \includegraphics[width=\textwidth]{"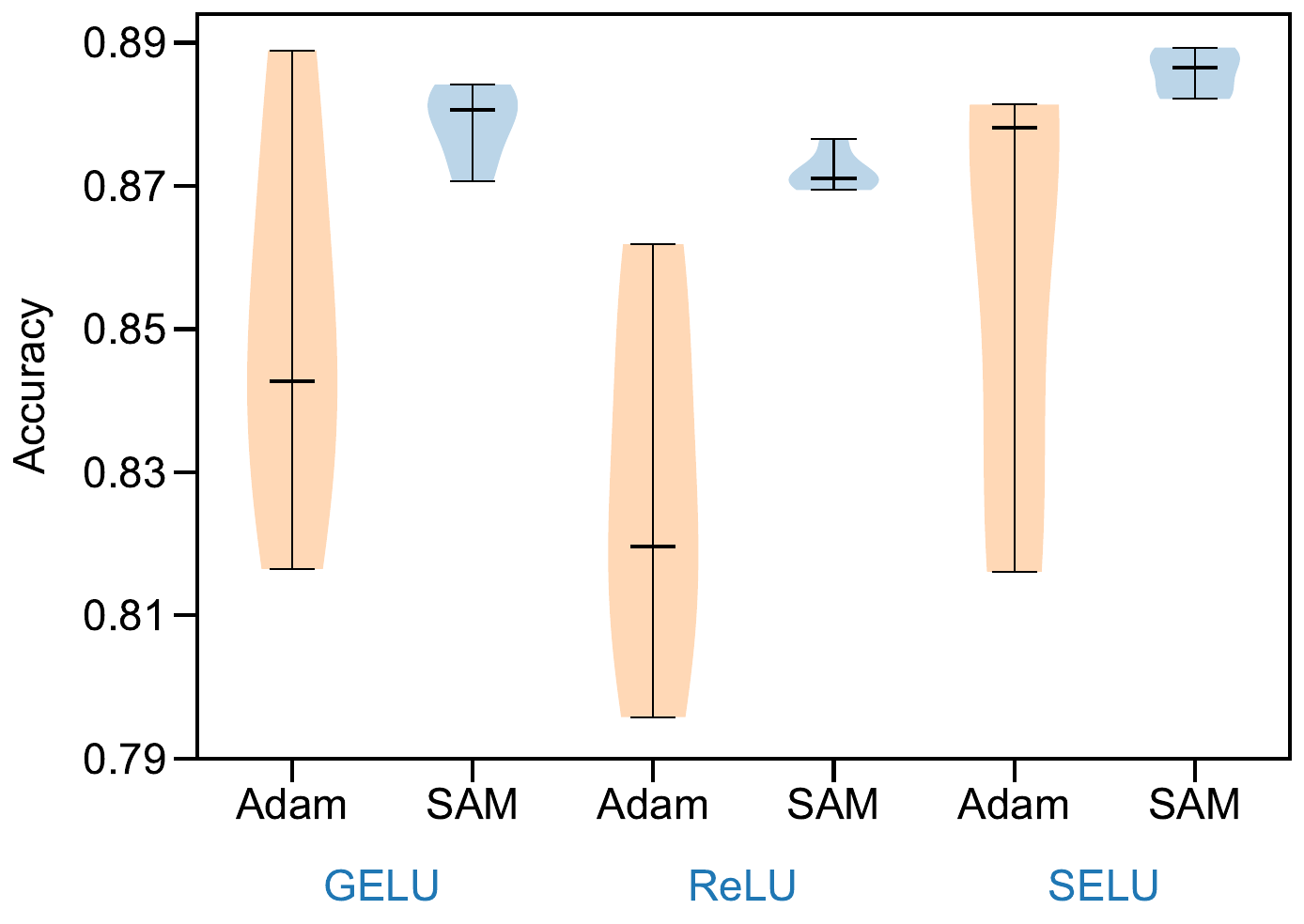"}
      \caption{Split 8}
    \end{subfigure}\hspace{\fill}%
    \begin{subfigure}{.25\textwidth}
      \centering
      \includegraphics[width=\textwidth]{"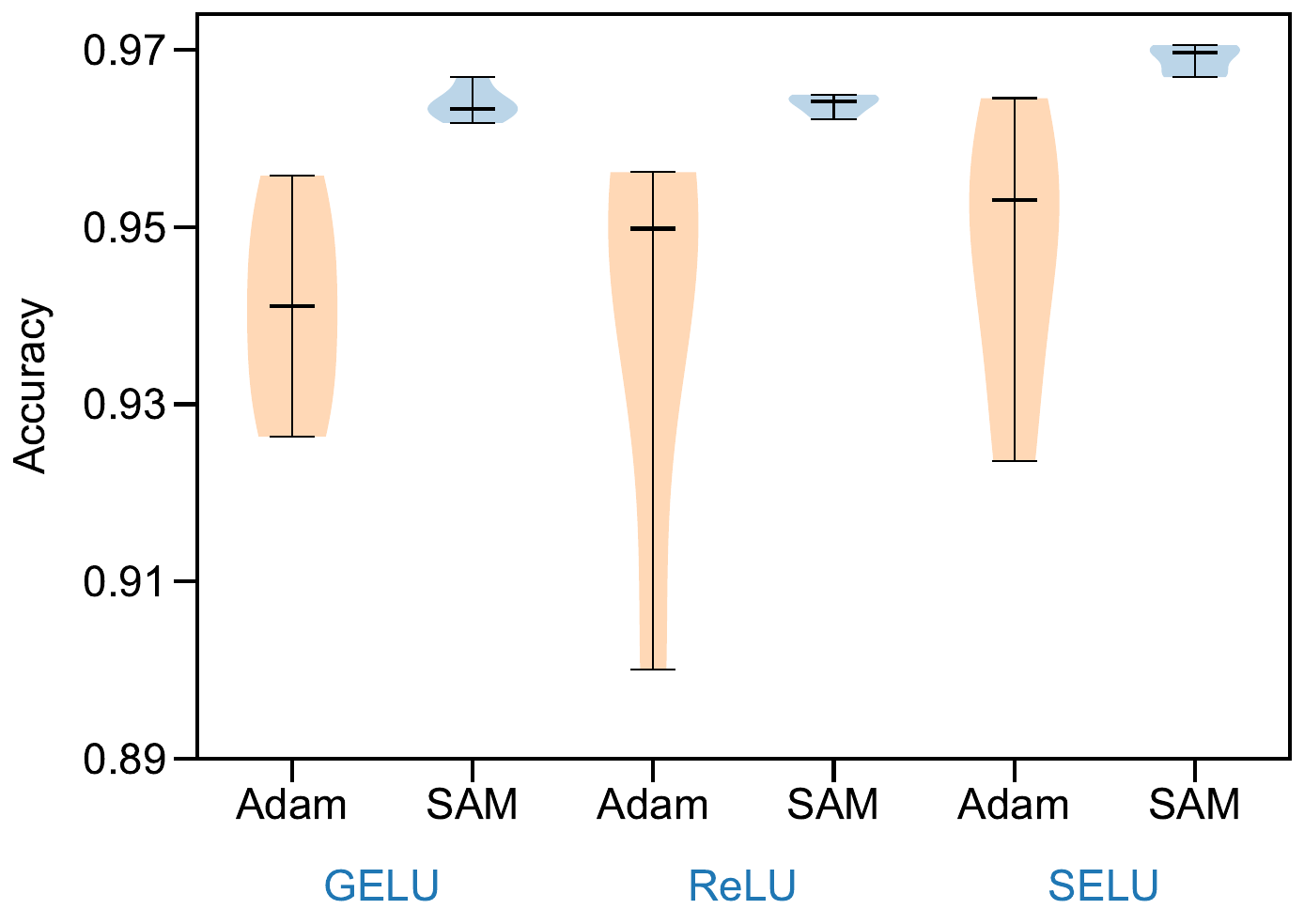"}
      \caption{Split 9}
    \end{subfigure}\hspace{\fill}%
    \bigskip
    \begin{subfigure}{.25\textwidth}
      \centering
      \includegraphics[width=\textwidth]{"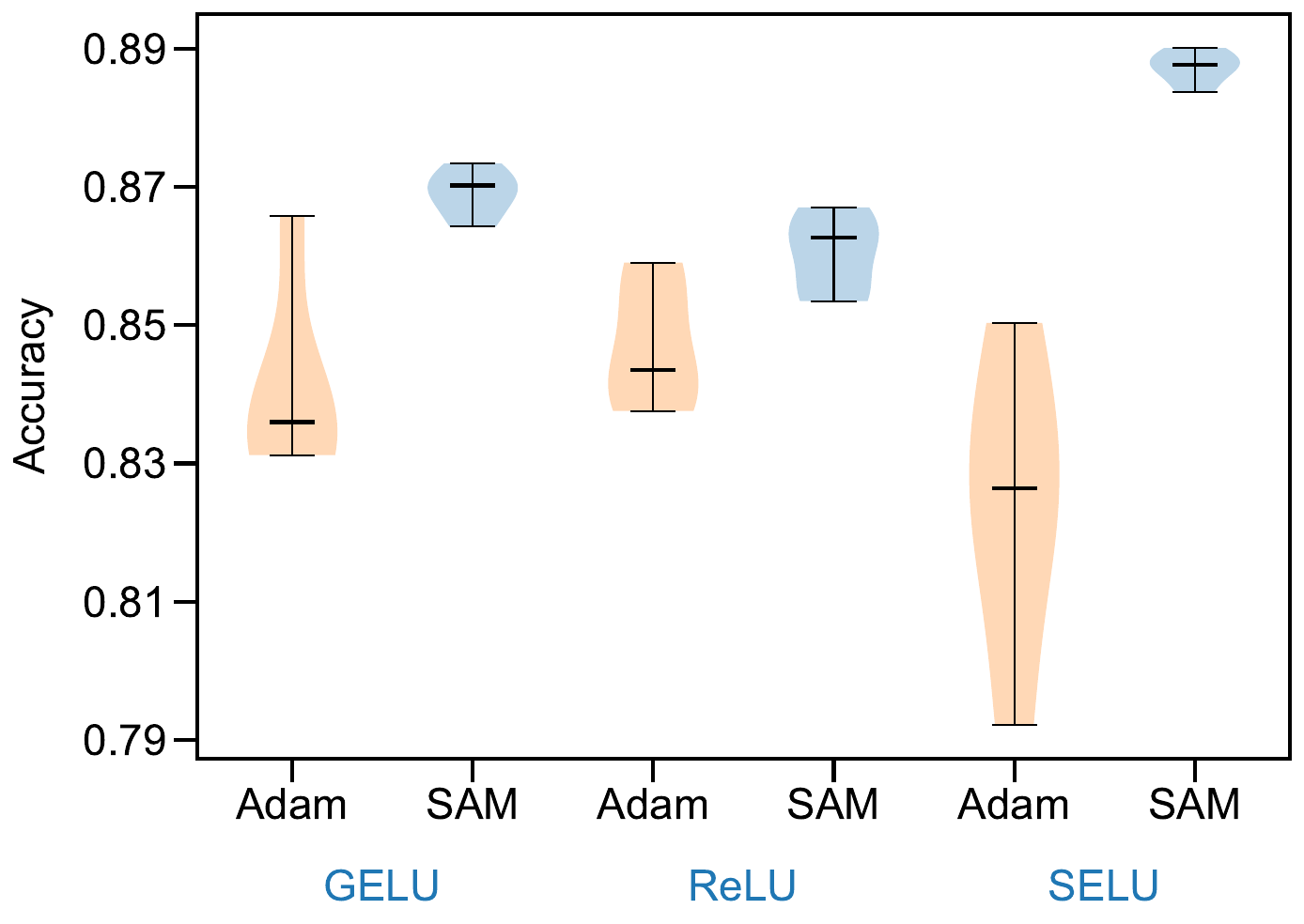"}
      \caption{Split 10}
    \end{subfigure}\hspace{\fill}%
    \caption{Individual plots showing classification performance $mean_{std}$ (n=5) for evaluation metrics on the testing set across the reported random splits.}
    \label{fig:appendix_2}
\end{figure*}

\end{document}